\RequirePackage{fix-cm}
\documentclass{article}
\usepackage[affil-it]{authblk}
\usepackage{graphicx}
\usepackage[utf8]{inputenc}
\usepackage{graphicx,url}
\usepackage{tikz}
\usepackage{verbatim}
\usepackage{tabularx}
\usepackage{dirtytalk}
\usepackage{tikz}
\usetikzlibrary{positioning}
\usetikzlibrary{arrows,calc,decorations.markings,math,arrows.meta}
\usetikzlibrary{graphs}
\usetikzlibrary{shapes,decorations}
\usepackage{hyperref}
\usepackage{booktabs}
\usepackage{setspace}
\usepackage{pgfplots}
\pgfplotsset{compat=1.16}
\usepackage[utf8]{inputenc}

\begin{document}

\title{An Update to the Minho Quotation Resource}


\author{Brett Drury} 
             \affil{LIAAD-INESC-Tec, Campus da FEUP \\  
             Rua Dr. Roberto Frias,  
             4200 - 465 Porto, Portugal \\
             brett.drury@gmail.com   }         
    
\author{    Samuel Morais Drury }
              \affil{Colégio Puríssimo ,R. Sete, 881 - Centro \\
              Rio Claro - SP \\
              Brazil}

\maketitle


\begin{abstract}
The Minho Quotation Resource was originally released in 2012. It provided approximately 500,000 quotes from business leaders, analysts and politicians that spanned the period from 2008 to 2012. The original resource had several failings which include a large number of missing job titles and affiliations as well as unnormalised job titles which produced a large variation in spellings and formats of the same employment position. Also, there were numerous duplicate posts. This update has standardised the job title text as well as the imputation of missing job titles and affiliations. Duplicate quotes have been deleted. This update also provides some metaphor and simile extraction as well as an emotion distribution of the quotes. This update has also replaced an antiquated version of Lucene index with a JSONL format as well as a rudimentary interface that can query the data supplied with the resource. It is hoped that this update will encourage the study of business communication in a time of a financial crisis. 
\end{abstract}

\section{Introduction}

The Minho Quotation Resource is a linguistic resource that contains 500,000 quotes from business and government leaders as well as analysts and economists that covers the period 2008 to 2012 which we released in 2012  \cite{drury2012minho}. The period covered by this resource is an economically interesting time because it is the start of the recovery from the financial recession of 2007, and contains quotes related to the economically significant events such as the bailouts of General Motors and Freddie Mae and Mac.

The original resource used \href{https://www.w3.org/2001/sw/wiki/Open_Calais}{Open Calais} to extract reported speech contained in business news. Open Calais returned a speaker name, and if available a job title and an organisation as well as a quote. An example of a complete relation is \textit{Ben Bernanke, Chairman, US Federal Reserve,\say{Considering some alternative forms for the government-sponsored enterprises - or for mortgage securitisation generally during this time out seems worthwhile.}.}. In this example, the Speaker is Ben Bernanke who held the post of Chairman at the US Federal Reserve.

In the original resource, there was no preprocessing of the data as well as no information extraction of significant words and phrases from the quote. 
The original release stored data in a Lucene Index and a CSV file. The resource now suffers from several issues and shortcomings. For example, the Lucene Index supplied with the resource is antiquated because it uses the Lucene format from 2012. In addition to the ageing format, there are some issues with the data itself such as missing job titles and affiliations as well as variations in speakers' names. And finally, the extractor has conflated affiliation with the job role.

These project notes describe a significant update of The Minho Quotation Resource. This update: imputes missing job titles and affiliations, metaphors and similes are extracted, the storage format is updated and an emotion distribution is estimated for each quote. 
 
These project notes will adhere to the following format: Corpus Characteristics, Phrase and Word Extraction, Resource Structure and Conclusion.

\section{Corpus Characteristics}

The original release of The Minho Quotation Resource was a selection of 500,000 quotes from business news, and it spans the period 2008 to 2012. It has several issues which are demonstrated by a simple frequency analysis which is shown in Table \ref{attributecount}. The frequency analysis counted the number of entries for the most frequent Speakers, Job Roles and Affiliation. The most frequent speaker is Barack Obama with 5979 attributed quotes. There are 131860 individual speakers in the resource. The most frequent Job Role is President with 9536 attributed quotes and the most frequent affiliation is Chief Executive with 5783 attributed quotes.

The frequency analysis demonstrates an error with the original resource. The error is the transposition of the Job Role with Affiliation. The underlying relation extraction has erroneously identified a job role as an affiliation which is shown by the most frequent affiliations. There are also a large number of speakers, job roles and affiliations which may be caused by non-normalised speaker name, job role and affiliation.

\begin{table}[!ht]
\begin{tabular}{lp{2.5cm}p{2.4cm}p{1.6cm}l}    \toprule
\emph{Attribute} &  \multicolumn{2}{l}{\emph{Most Frequent}} &  \multicolumn{2}{l}{\emph{Number of Entries}} \\\midrule
 Speaker    & B.Obama   & G.Brown  & B.Bernanke & 131860 \\
 & 5979 & 1783 & 1498 & \\
 \hline
 Job Role & President  & Chief Executive & CEO & 14050 \\
 &  9536 & 9064 & 8131 & \\
 \hline
 Affiliation & Chief Executive & President & Minister & 65004 \\
 & 5783 & 3362 & 2073 & \\
 \hline
\end{tabular}
\caption{Count of Attribute Entries}
\label{attributecount}
\end{table}

\subsection{Corpus Preprocessing}

It is clear from the previous section that a preprocessing step was required to remove the transposition error, reduce speaker name and job title ambiguity, remove duplicated and non-English sentences, and impute missing job titles and affiliations. 

The transposition error was resolved by matching the entries in the affiliation category with the entries in the job title category and if they matched the affiliation term was moved to the job title category. For example, in Table \ref{attributecount} it is clear that Chief Executive is present in both the Affiliation and Job Role category, consequently any Chief Executive entry in the Affiliation category would be moved to the Job Role category.

The speaker name ambiguity was reduced by changing all names to title case. Therefore the entry \say{barack Obama} would be changed to \say{Barack Obama}. Manual rules changed entries of people who are known by a single name to their complete name. For example, the entry \say{Obama} was expanded to \say{Barack Obama}. 

The final step in name normalisation was to expand single names to similar full names who have the same affiliation. These individuals are not known by a single name. For example, Dick Fuld who was the CEO of Lehman Brothers was often referred to as Fuld. If the Quote where he was referred to as Fuld had his affiliation as Lehman Brothers, then his name would be expanded to Dick Fuld. If the quote had his affiliation missing then his name would remain the same. This is the case for the Quote with the ID 341819.  

The job title ambiguity was reduced by contracting three-word job titles to an acronym such as \say{Chief Operating Office} to \say{COO}. Also, there were manual rules that transformed known job titles that had less than three words to an acronym. For example, the job role of Chief Executive would be transformed to CEO. Compound job titles were simplified to one job title by splitting the job title at the conjunction and selecting the job title on the left-hand side. For example, the job title \say{President and CEO} would be simplified to \say{President}.  

Missing job titles were imputed by matching a candidate name and affiliation with an entry that has the corresponding name and affiliation as well as a job title. The job title is then transferred to the entry with the candidate name and affiliation. For example, if there was an entry for \say{Dick Fuld} with the affiliation \say{Lehman Brothers}, but no job role, then he would be assigned to the job title \say{CEO} because the majority of Quotes which had the name \say{Dick Fuld} and affiliation \say{Lehman Brothers} had a job role of \say{CEO}. 

The same process was followed for missing affiliations, but the matching was by using name and job title rather than name and affiliation. For example, using the \say{Dick Fuld} example, an entry that had \say{Dick Fuld} and the job role \say{CEO} would be assigned the affiliation \say{Lehman Brothers} because the majority of entries the name \say{Dick Fuld} and the job role \say{CEO} have the affiliation \say{Lehman Brothers}.  

Some entries were duplicated because of the repetition of quotes reported in different media sources. The removal duplicate quotes which was achieved by matching speaker name and quotes. If more there was more than one quote with the same speaker name, then one quote would be removed. In the much used Dick Fuld example, if there were more than one entry with the following details: \textit{Dick Fuld    CEO    Lehman Brothers     You know what, let them line up; I can handle it.}, one of them would be deleted.

There were quotes not reported in English. These quotes were identified by detecting the language of the quote with the \textit{LangDetect} Python Library. Any quotes identified by the library as non-English were removed.

The final step of the preprocessing stage was the unescaping of HTML entities and the removal of HTML from the quotes. Text which is marked up in HTML often escaped and the escaped text is then rendered by the browser. For example, \textit{\&lt;} is the escaped form of \textit{<}. The unescaping returned the escaped character to its original form. Therefore the quote text is represented uniformly. Any remaining HTML was removed with the \textit{Beautiful Soup} Python Library as well as manual rules. Any quotes which were empty of this process were removed.

The results of the preprocessing are shown in Table \ref{refinedattributecount}, and it is clear from the Table that the amount of different total entries for each attribute have been reduced, and that the most frequent affiliations now match that of the most frequent speakers.  

\begin{table}[!ht]
\begin{tabular}{lp{2.5cm}p{2.5cm}p{1.8cm}l}    \toprule
\emph{Attribute} &  \multicolumn{2}{l}{\emph{Most Frequent}} &  \multicolumn{2}{l}{\emph{Number of Entries}} \\\midrule
Speaker    & B.Obama   & G.Brown  & B.Bernanke &  130942 \\
 & 6524 & 1685 & 1256 &  \\
 \hline
Job Role & CEO  & President & Chairman & 5358\\
& 69175 & 38155 & 18427 &  \\
\hline
Affiliation & White House  & UK Government  &  US Federal Reserve & 50256 \\ 
& 7996 & 2841 & 2229 & \\
\hline

\end{tabular}
\caption{Count of Attribute Entries After Preprocessing}
\label{refinedattributecount}
\end{table}

The imputation of missing job titles and affiliations reduced the number of missing entries in the aforementioned columns. This is demonstrated in Table \ref{blankcount}, which shows the number of missing entries before and after the imputation step. The number of missing entries for Job Role and Affiliation has been reduced significantly. The missing entries in the job role category have been reduced by  71.40\% (378114 to 108138) and the affiliation category by  82.90\% (293485 to 50256).   

The preprocessing phase has reduced the amount of ambiguity and the number of missing entries and after preprocessing the number of quotes have been reduced to 458602 entries due to the removal of duplicate, blank and non-English quotes.   

\begin{table}[!ht]
\begin{tabular}{lp{2.5cm}p{2.5cm}p{1.8cm}l}    \toprule
\emph{Attribute} &  \multicolumn{2}{l}{\emph{Blank Entries (Unprocessed)}} &  \multicolumn{2}{l}{\emph{Blank Entries (Processed)}} \\\midrule
Speaker & \multicolumn{2}{l}{0} & \multicolumn{2}{l}{0} \\
\hline
Job Role & \multicolumn{2}{l}{378114} & \multicolumn{2}{l}{108138} \\
\hline
Affiliation  & \multicolumn{2}{l}{293485} & \multicolumn{2}{l}{50256} \\
\hline
\end{tabular}
\caption{Count of Blank Entries}
\label{blankcount}
\end{table}

\section{Phrase and Word Extraction}  

Phrases and words from the quote may carry information that may assist the interpretation of the quote. The word and phrase types that were extracted from each quote were: highlighted terms, neologisms, similes and metaphors.

Highlighted terms are words and phrases that have been given an indicator of importance or derision by the journalist by using inverted commas. For example, \say{A \$10bn credit line that it drew down earlier this year is "needed for VW" and is not invested in options.}, the highlighted term is \say{needed for VW}, and it can be inferred from the quote that the journalist is derisive of the claim by Frank Gaube that VW needs a \$10Bn credit line is needed by VW \cite{FT}. These terms were extracted using manual rules.

Neologisms are invented words. Neologisms are arguably an attempt by the speaker to convey a meaning that is not available in the English lexicon. For example, \say{The \textbf{measurability} really sets it apart.; You can track any money you spend online} \cite{indie}. In this case, the neologism is the term \textbf{measurability} which is a portmanteau of the words \say{measure} and \say{ability}, which conveys the ease a process has to be measured. The neologisms were detected by identifying words that are not in English Language dictionaries, and are not a named entity such as company or person name as well the word not being a typo, spelling error or formatting error. The neologism identification step checked the candidate word against US and UK English dictionaries, as well as checking the word against the speaker's name, job role and affiliation as well as against named entities extracted from the quote by a pre-trained model. If the candidate word fails each one of these checks it is assumed to be a candidate neologism. Spelling and formatting errors are migrated by ensuring that the candidate neologism appears in the corpus a minimum number of times.
\subsection{Simile Extraction}

A simile is a figure of speech that compares one thing with a different thing using the connectors \say{like a} or \say{as a}. The use of similes in business communication is an aid to communication with a wide audience because it can make \say{an arid topic concrete
and immediate} \cite{kallendorf1985figures}. And therefore the communicator will not lose the attention of the audience. Similes can often be crude such as the example provided by \cite{kallendorf1985figures} which drew a comparison between sex and labour relations

The importance of similes in business communication was the motivation to extract it and provide it as part of the resource. The extraction technique was a rule-based approach which used the following patterns to extract similes.  
\begin{enumerate}
    \item \textbf{ADJ}$_{0:1}$\textbf{NOUN}$_1$\textbf{like}$_1$\textbf{DET}$_{0:1}$(\textbf{ADJ}$_{0:1}$\textbf{NOUN}$_1$)$_{1:}$
    
    \item \textbf{as}$_{1}$\textbf{ADJ}$_{1}$\textbf{as}$_{1}$\textbf{DET}$_{0:1}$(\textbf{ADJ}$_{0:1}$\textbf{NOUN}$_1$)$_{1:}$
\end{enumerate}

The aforementioned Rules uses the following nomenclature: Noun: Noun, ADJ: Adjective and DET: Determiner. The numeric arguments determine the minimum and maximum times the term can appear in the extracted phrase. For example, $_{0:1}$ indicates that the minimum amount of times it appears is $0$ and the maximum frequency is $1$, whereas $1$ indicates that the term must appear exactly once. Where is there is no upper limit the term can appear as many times as it exists in the quote.

The first rule identifies similes that compare Nouns or Adjective-Noun pairs with the term \say{like} with other Nouns or Adjective-Noun Pairs. The rule also allows for the optional use of determiners. Rule one for example returns this simile: \textit{currency like the dollar}, where the word \textit{currency} is the Left Hand Side (LHS) Noun, and \textit{the} is the optimal determiner and the word \textit{dollar} is the the Right Hand Side (RHS) Noun. The simile \textit{transformational technologies like broadband} which is also returned by Rule one, demonstrates the use of the optional adjective \textit{transformational}.

The second rule identifies similes that compare an Adjective with Nouns or Adjective-Noun pairs with the phrase \textit{as a}. In common with Rule One, Rule Two also allows for the use of optional determiners. An example returned by Rule Two is \textit{as good as a coin flip} where the adjective \textit{good} is compared with the noun phrase \say{coin flip}. This phrase does not have an optional determiner. 

\subsection{Metaphor Extraction}

A metaphor is a word or phrase that describes an action that is applied to an object which is not physically possible, such as \textit{leap of faith}. The role of metaphor in business and financial language is to illustrate complex or dry language with visual stories that indicate the thought process of the speaker \cite{vasiloaia2011metaphors}. The resource has two types of metaphors extracted for each quote. They are: \say{is a} and \say{of} metaphors. 

The \say{is a} metaphor extraction strategy follows the technique described by \cite{krishnakumaran2007hunting} where the metaphor is detected by the following pattern: \textit{NP is a NP}, where $NP$ represents a Noun Phrase. The last step of the technique is to ensure that the nouns in the Noun Phrases are not hyponyms of the other and that they do not share any common hyponyms.  An example of a \say{is a} metaphor returned by this technique is \textit{This tax credit is a lifeline}.

The \say{of} metaphor was computed by the technique described by \cite{pmi}. With this technique candidate pairs of Noun and Noun Phrases are harvested with the pattern: \textit{NP of NP} where $NP$ represents a Noun Phrase. A pointwise mutual information score is computed by: $P(x,y) = log \frac{p(y|x)}{p(x)}$, where $x$ is the RHS Noun Phrase and $y$ is the LHS Noun Phrase, and $p(y|x)$ is the probability of two Noun Phrases co-occurring whereas  $p(x)$ is the probability of the Noun Phrase $x$ appearing in the whole corpus.  Noun Phrase pairs that are above a predetermined score are accepted as constituents of a  \say{of} metaphor such as \say{leap of faith} and \say{ghettos of poverty}.

\subsection{Emotion Detection}
Emotion detection is a technique of classifying a word or phrase into an emotion category such as joy or anger.
The motivation to add this information is that emotions drive the financial markets and can be an indicator of the underlying prospects of the company \cite{strauss2016lagging}. The tool that was used is called \href{https://github.com/glhuilli/limbic}{Limbic} which classifies a quote into one of the four following emotion categories:  anger, fear, joy and sadness. The score for each category is a continuous number from zero to one. With a score of one being the highest score possible. The emotion category the quote is classified into is the emotion with the highest score.

The quotes that attracted high scores for each emotion is shown in Table \ref{Emotive}. The emotions estimated from the economic actors represented in this corpus indicates Keynes \say{Animal Spirits}\cite{keynes1937general} stalked the financial markets between 2008 and 2012.

\begin{table}[!ht]
    \centering
    \begin{tabular}{c|c|p{8.5cm}}
       \hline
       Emotion & Score & Quote \\
       \hline
        Anger & 0.92 &  The technology industry would be likely to fare better this time than during its last previous big downturn in the aftermath of the tech bubble in 2001, partly owing to advances in how inventory is managed. "We all feel the rage of the storm" \\ 
        \hline
        Fear & 0.84 & Some analysts attribute the correlation to continuing foreign capital outflows, while others say it is panic among domestic investors. "It's all Gulf sentiment now," \\
        \hline
        Joy & 0.89 & the forecast fall in retention numbers was a prudent step. "Our customers tend to be elderly, insurance-minded people who are happy to pay to avoid the worry of a household emergency," \\
        \hline
        Sadness & 0.90 & "While risk appetite may be getting a bit frothy ... the pound may remain resilient versus the euro with the still festering problems for the eurozone largely left unaddressed, especially the terrible costs that the strong euro is having on its weaker members. \\
        \hline
         
    \end{tabular}
    \caption{Emotive Quotations}
    \label{Emotive}
\end{table}

\subsection{Entity Detection}
A pre-trained entity detection model was used to identify named entities present in the quote. The full list of named entity classes that are detected can be found \href{https://spacy.io/api/annotation#named-entities}{here}, and include Person, Location and Product. Sample output is the following: "Eric", "PERSON", "Larry", "PERSON", "Natti", "PERSON", which was generated from the quote: \say{Eric, Larry, and Natti are some of the more talented equity sales and trading professionals on the Street}.
\section{Resource Structure}

The updated resource is located \href{https://github.com/bmd123/MinhoQuotationResource}{here}. The resource contains the following data files: Raw Data, Sentiment Bigrams, Discourse Connectors and the JSONL file which contains the updated corpus. 

The Raw Data file is a tab-delimited file (TSV) that contains a Quote ID, a Speaker Name, an Affiliation and a Quote. The Sentiment Bigrams and Discourse Connectors were created for the first release of this resource \cite{drury2012minho}. The discourse connectors contain a discourse connector and a sentiment score, whereas the sentiment bigrams are word pair candidates with a sentiment score computed by the method described by \cite{liu2007web}. Each file has a single entry per line with an unbounded number where a score of less than zero indicates a negative sentiment and a score above zero indicates a positive score. A representative entry for a discourse entry is \say{worried} which has a score of -2.49. The bigram \say{strong reputation} which has a score of 16.10 is a typical entry for the sentiment bigrams. 

The JSONL file contains markup that describes a Python dictionary for each quote. It contains the following keys: 

\begin{itemize}
    \item ID: {\textit{Unique identifier for the entry.}}
    \item Speaker Name:  {\textit{The name of the person who made the quote.}}
    \item Affiliation: {\textit{The organisation that the speaker is associated with.}}
    \item Sentences: {\textit{The sentences of quote}}
    \item Neologism: {\textit{Neologisms present in the quote }}
    \item Quotedphrases: {\textit{Word and phrases in the quote that have been highlighted by the journalist }}
    \item Simile: {\textit{Similes present in the quote}}
    \item Metaphor: {\textit{Metaphors present in the quote}}
    \item Emotion: {\textit{Emotion distribution of the quote}}
    \item Entities: {\textit{Entities present in the quote}}
\end{itemize}

The resource contains one code file, \say{main.py} which provides a rudimentary interface to the JSONL file.  The user passes a command-line argument of a dictionary key with a comma-delimited of words that will be searched for as well as an output location of an image that a Word Cloud will be outputted to.

\section{Conclusion}

These Project Notes describe an update to the Minho Quotation Resource. This update includes removal of duplicate quotes, normalisation of job titles as well as imputation of job roles and affiliations. In addition to the resources supplied with the first iteration, metaphors and similes, as well as emotion distribution for each quote, have been added. 

The resource covers an important period of financial history where the world suffered a depression. The resource contains quotes about some significant events such as the failure of Lehman Brothers and the bailout of General Motors as well as quotes from important economic actors such as Ben Bernanke and Richard Fuld.  As the world economy begins to enter an economic crisis this update is a timely release for scholars who wish to study how economic actors communicate in a period of a severe financial crisis.  

\bibliographystyle{plain}
\bibliography{bib}
\end{document}